\pdfoutput=1
\documentclass[11pt,a4paper]{article}
\usepackage[hyperref]{emnlp2020}
\usepackage{times}
\usepackage{latexsym}
\usepackage{multirow}
\usepackage{pifont}

\usepackage{microtype}

\aclfinalcopy 

\title{PublishInCovid19 at WNUT 2020 Shared Task-1: Entity Recognition in Wet Lab Protocols using Structured Learning Ensemble and Contextualised Embeddings}

\author{Janvijay Singh and Anshul Wadhawan\\
  Flipkart Private Limited \\
  \texttt{\{janvijay.singh,anshul.wadhwan\}@flipkart.com} \\}

\date{}

\begin{document}
\maketitle
\begin{abstract}
In this paper, we describe the approach that we employed to address the task of Entity Recognition over Wet Lab Protocols - a shared task in EMNLP WNUT-2020 Workshop. Our approach is composed of two phases. In the first phase, we experiment with various contextualised word embeddings (like Flair, BERT-based) and a BiLSTM-CRF model to arrive at the best-performing architecture. In the second phase, we create an ensemble composed of eleven BiLSTM-CRF models. The individual models are trained on random train-validation splits of the complete dataset. Here, we also experiment with different output merging schemes, including Majority Voting and Structured Learning Ensembling (SLE). Our final submission achieved a micro F1-score of 0.8175 and 0.7757 for the partial and exact match of the entity spans, respectively. We were ranked first and second, in terms of partial and exact match, respectively.
\end{abstract}

\section{Introduction}

Entity Recognition (aka entity extraction or chunking) involves detection (begin and end boundaries) and classification of entities mentioned in unstructured text into pre-defined categories. It is one of the foundational sub-task of several Information Extraction \cite{10.1007/978-3-319-05476-6_12} (IE) and Natural Language Processing (NLP) pipelines. Hence, errors introduced during the extraction of entities can propagate further and degrade the performance of the complete IE or NLP pipeline. In the domains of experimental biology, the growing complexity of experiments has resulted in a need to automate wet laboratory procedures. Such an automation will be useful in avoiding human errors introduced in the wet lab protocols and thereby will enhance the reproducibility of experimental biological research. 

To achieve this reproducibility, some of the previous research works have focussed on defining machine-readable formats for writing wet lab protocols \citep{king2009automation,ananthanarayanan2010biocoder,vasilev}. However, the vast majority of today’s protocols are written in natural language with jargon and colloquial language constructs that emerge as a byproduct of ad-hoc protocol documentation. This motivates the need for machine reading systems that can interpret the meaning of these natural language instructions, to enhance reproducibility via semantic protocols (e.g. the Aquarium project) and enable robotic automation \citep{bates2017wet} by mapping natural language instructions to executable actions. In order to enable research on interpreting natural language instructions, with practical applications in biology and life sciences, an annotated database \citep{kulkarni2018wetlab} of wet lab protocols was introduced. 

The first step in interpreting natural language lab protocols is to extract entities, followed by identification of relations between them. To address the research focussing on entity recognition over Wet Lab Protocols a shared task \citep{tabassum2020wlp} was introduced at EMNLP WNUT-2020 Workshop. The task was based on the annotated database \citep{kulkarni2018wetlab} of wet lab protocols. We tackle this task in two phases. In the first phase, we experiment with various contextualised word embeddings (like Flair, BERT-based) and a BiLSTM-CRF model to arrive at the best-performing architecture. In the second phase, we create an ensemble composed of eleven BiLSTM-CRF models. The individual models are trained on random train-validation splits of the complete dataset. Here, we also experiment with different output merging schemes, including Majority Voting and SLE.

The rest of the paper is structured as follows: Section 2 states the task definition. Section 3 describes the specifics of our methodology. Section 4 explains the experimental setup and the results, and Section 5 concludes the paper.

\section{Task Definition}

The steps involved in any lab procedure are specified by lab protocols. These protocols have several characteristics like noise, density and domain specificity. Any process that can automatically or semi-automatically convert protocols into a format that machine recognizes advantages biological research. In this task, system entries for entity recognition on a dataset of lab protocols are invited. Since the protocols are written manually by lab technicians and researchers, they are subject to spelling errors and non standard language.  

The data provided in the task is made available in two formats:

\subsection{CoNLL format} In this format, each line represents the named entity in the following manner:

\begin{center}
\textless word \textgreater + "\textbackslash t"+ \textless NE \textgreater
\end{center}

An empty line denotes the end of a sentence.

\subsection{Standoff format} The standoff format contains each protocol represented by two separate files. One file, with .txt extension, contains protocols in text format, while the other file, with .ann extension, contains protocol annotations. The two files are linked by using a simple file naming convention wherein their base name is the same, i.e. the file name without the extension is the same. For example,the annotation file named as protocol\_17.ann contains annotations for the file protocol\_17.txt.

Within each annotation file, individual annotations connect to different parts of text through character offsets. For example, in the document starting as “Put 3.68 g of NaCl”, the text “Put” is denoted by the offset range 0..3. It is evident from the above example that all offsets are 0 indexed and include the character at the start offset and exclude the character at the end offset. All text files have the file extension .txt and contain the text of original documents provided as inputs to the system. The encoding used in the protocol text files which are stored as plain text files is UTF-8 (an extension of ASCII). Each line in the protocol text file denotes a single step in the protocol. Hence, all steps in the entire protocol are separated by newline characters. The first line in every file indicates the protocol’s name/title.

\section{Methodology}
This section talks about the core methodology we adopted to tackle the given problem. The process pipeline involves providing contextualised word embeddings as input to the BiLSTM-CRF model, followed by a Structured learning Ensemble approach. Each of the these modules have been described in detail in the below subsections.

\subsection{Embeddings}

We experiment with two types of contextualised word embeddings, BERT and Flair based, which we discuss in detail in the below subsections.

\subsubsection{BERT}

Neural models based on transformers  \cite{vaswani2017attention} have excelled in most NLP tasks. The primary components in their architecture being the self attention blocks and feed forward layers, these models have been proven successful in providing a significant boost to state-of-the-art results. The major difference between transformers and RNN based models \cite{li2018survey} is that transformers do not rely on recurrence mechanisms to establish relations and dependencies in the input sequence, by making use of self attention at each input time step instead. Attention can be interpreted as a technique to map a query and a set of key-value pairs to an output, where the query, keys, values and output are all vectors. As far as self attention is concerned, a separate feed forward layer is used to formulate the query, key and value vectors for each vector in the input sequence. For every input vector, the score for attention is calculated using a compatibility function which takes as input the input keys and query vector. These attention scores are used to denote the weights of a weighted sum of value vectors, which is the output of self attention technique. Another technique widely used is the multi headed attention technique in which several modules of these self attention blocks work over the input sequence. The encoder module in the transformer’s architecture contains 6 identical layers each having two sublayers - position wise densely connected feed forward network and multi headed self attention layers. These sublayers are wrapped around with residual connections. Layer normalisation follows the above module. BERT pre-trains bidirectional representations by jointly utilizing both right and left contexts across all layers with the help of a multi layer encoder module. These pre-trained BERT representations are then fine tuned as per the required task by appending a separate output layer depending on the task to be performed.   

For every token, the summation of the corresponding token, segment and position embeddings is carried  out to produce BERT’s input representation. The training process for BERT involves Masked Language Modelling \cite{nozza2020mask} and Next Sentence Prediction \cite{shi-demberg-2019-next}, both of which are unsupervised prediction tasks. BERT representation for each token in the input text is then fed to the appended densely connected layers to produce the output labels for the token as part of the fine tuning process. The predictions produced are independent of the surrounding predictions produced. 

We experimented with different variations of BERT models \cite{devlin2018bert} for generating word embeddings. All the listed model types have 12 layers, 12 attention heads and 110M parameters.

{\bf BERT-base-cased} : This model is trained on cased English text of general domain like Wikipedia text and BooksCorpus.


{\bf BioBERT } \cite{Lee_2019} : BioBERT is a language representation model pre-trained on the domain of biomedical data. The pre-training process for BioBERT involves initializing weights with those of BERT which is pre-trained on general domain corpora, followed by pre-training BioBERT with biomedical data corpora like PMC full-text articles and PubMed abstracts. 

{\bf PubMedBERT} \cite{pubmedbert} : The base architecture of PubMedBERT is the same as an uncased BERT base model. The model is pre-trained on full PubMed Central articles and PubMed abstracts. The pre-training process for this model involves direct pre-training on biomedical text from scratch. Thus, the weights are not initialized with those of BERT as was in the case of BioBERT. The pre-training corpus contains 14 million PubMed abstracts with 3 billion words, 21 GB of textual data in total. Another version of the same model is pre-trained on additional data of full text PubMed Central articles, with the total textual data containing 16.8 billion words and 107 GB in size.

\subsubsection{Flair}

\footnote{\url{https://github.com/flairNLP/flair}} Flair embeddings are pre-trained Contextualised Word Embeddings (CWE) provided in the Flair NLP framework. In contrast to classical work embeddings like GloVe, the Flair CWE concatenate two context vectors based on the left and right sentence context of the word to it. These context vectors are computed using two recurrent neural models. One of the character language model is trained from left to right while the other is trained from right to left. Flair CWEs have been applied successfully to sequence tagging tasks such as Named Entity Recognition and Part of Speech Tagging. Since this shared task is closely related to Bio-medical domain, we have used “pubmed” variant of Flair CWEs in all our experiments.

\subsection{BiLSTM-CRF Model}

The ability of Recurrent Neural Networks (RNNs) \cite{yadav2019survey} to execute the same function at each time step, allowing parameters to be shared across the input sequence, make them highly suitable for sequential input data . Useful information from each time step is forwarded to further time steps in the form of a hidden vector, which is utilized to make a prediction at each of the future steps. However, RNNs face the issue of vanishing gradients in case of large input sequences. To solve this issue of vanishing gradients, (Long Short Term Memory) LSTM \cite{Hochreiter1997LongSM} was introduced. The presence of gating mechanisms in LSTMs makes sure that long range dependencies are captured appropriately. While LSTMs utilize only past time steps to make a prediction, Bidirectional LSTM (BiLSTM) \cite{schuster1997bidirectional} utilizes information from past as well as future time steps. In our case, the output embeddings are fed to the BiLSTM layer, which outputs a vector for each word in the input sequence. Since the task under consideration has labels which have dependencies among themselves, such as an intermediate\_label following a start\_label, we need to consider these dependencies in our modelling approach. For this, a linear chain (Conditional Random Fields) CRF layer \cite{sutton2010introduction} is appended to the BiLSTM layer. Due to utilization of transition matrices for output labels, a linear chain CRF is able to learn inter label dependencies, if any, among the output labels.  

\subsection{Ensemble Process}

We created eleven randomly shuffled splits of training and validation data, and fine tuned our final model on these eleven splits to produce eleven sets of predictions. We then merged these predictions following two merging techniques, Majority Voting and Structured Learning Ensemble (SLE), thus comparing the performance of the two merging functions. In our experiments, we provide a fair comparison of the above two combination techniques, i.e. Majority Voting technique and SLE.

Given N number of ensembles and x as the input example, \{y\textsubscript{1} , y\textsubscript{2} , ..., y\textsubscript{N}\} being the predictions from N different models are merged to produce the final prediction y. The ensemble methods for structured output classification and multiclass classification differ in the way they merge the predicted results of the base models.  

The merging techniques have been described below:  

\subsubsection{Majority voting}

For every entity predicted, we choose the mode i.e. the most frequently occurring entity among the eleven predictions \cite{article}. Thus, the entity which has the maximum number of votes wins.  

Mathematically, the above process of majority voting scheme to produce the final predictions can be denoted in the below manner :\newline 

$ \mathbf{y} = \langle \left.{majority} \left\{ \left(\mathbf{y}_{1}\right)_{1},\left(\mathbf{y}_{2}\right)_{1},\ldots,\left(\mathbf{y}_{N}\right)_{1} \right\}\right\rangle $
\ldots \ldots \ldots .
$ \left.{majority} \left\{ \left(\mathbf{y}_{1}\right)_{L},\left(\mathbf{y}_{2}\right)_{L},\ldots,\left(\mathbf{y}_{N}\right)_{L} \right\}\right\rangle \newline$


where L is the length of all predictions. 

\subsubsection{Structured Learning Ensemble (SLE)}

Due to the presence of correlations and intrinsic structures in the output labels, we speculated that the majority voting scheme would not suffice for our problem. \cite{10.1145/1273496.1273582} proposed a technique to combine the predictions considering the correlations of the output labels. Named as weighted transition combination, the algorithm involves construction of (L-1) transition matrices of size ( \begin{math} \vert \Sigma \vert \end{math} x \begin{math} \vert \Sigma \vert \end{math} ) , where \begin{math} \Sigma \end{math} is the set of all possible labels. Apart from this, it also involves construction of a transition matrix T\textsuperscript{k} which provides the number of transitions at the k\textsuperscript{th} position as follows:

\[ T^{k} \left (t_{i}, t_{j} \right )=  {count}_{k} \left (t_{i}, t_{j} \right), \forall 1 \leq k \leq (L-1) \]

where count\textsubscript{k}(t\textsubscript{i}, t\textsubscript{j}) denotes the number of times the label t\textsubscript{j} occurs after t\textsubscript{i} at the k\textsuperscript{th} position in the set of predicted sequences \{y\textsubscript{1} , y\textsubscript{2} , ..., y\textsubscript{N}\}. Also, a stateweight vector is constructed that denotes the number of times label t\textsubscript{i} occurs at position k in the predicted sequences.

\[ U^{k}\left(t_{i}\right)={count}_{k}\left(t_{i}\right), \forall 1 \leq k \leq L \]

The predicted sequence of SLE is given by:

\[ \mathbf{y} = {argmax}_{\mathbf{y}} \prod_{k=1}^{L-1} T^{k}\left(y_{k}, y_{k+1}\right) \prod_{k=1}^{L} U^{k}\left(y_{k}\right) \]

The computation involved in the argmax calculation of the above equation is similar to Viterbi dynamic programming approach. 

\section{Experiments}

Our experimentation strategy is distributed in two phases. In the first phase, we experiment with various architectures and their specifications by varying the type of pre-trained model, deciding layers to freeze i.e. complete fine-tuning or contextual word embeddings, varying type and size of final layer in order to arrive at the best performing model. We trained each of our model architectures on the train split and identified the checkpoint which worked best using the validation split. We reported the final numbers on the test split. For each model, we train three different models with random seed values and then report averaged f1 scores to ensure that improvements are not the result of randomisation. A configuration of concatenated contextual word embeddings from PubmedBERT and Flair, followed by 2 BiLSTM layers with 512 dimensional hidden size and a CRF layer in the end worked best. In the second phase, we train individual models on random splits of train + validation sets. In order to merge the outputs of individual models, we experiment with two output merging schemes namely Majority Voting and Structured Learning Ensemble (SLE). Finally, we report the results on the test dataset.

In the following sub-sections, we describe the dataset, system settings, evaluation metrics, results and a brief error analysis for our final submitted system.

\begin{table*}[]
\centering
\begin{tabular}{|l|l|l|l|l|}
\hline
 & train\_data & dev\_data & test\_data & test\_data\_2020 \\ \hline
Measure-Type & 857 & 329 & 272 & 731 \\ \hline
Numerical & 838 & 262 & 231 & 520 \\ \hline
Size & 262 & 124 & 114 & 238 \\ \hline
Seal & 210 & 92 & 64 & 119 \\ \hline
Speed & 626 & 241 & 167 & 240 \\ \hline
Location & 3929 & 1407 & 1327 & 1670 \\ \hline
Temperature & 1594 & 492 & 532 & 760 \\ \hline
Amount & 3438 & 1102 & 1193 & 1238 \\ \hline
Method & 1605 & 545 & 582 & 1077 \\ \hline
pH & 67 & 37 & 62 & 66 \\ \hline
Generic-Measure & 487 & 136 & 143 & 176 \\ \hline
O & 53690 & 18454 & 17925 & 26012 \\ \hline
Action & 12368 & 4057 & 4140 & 5439 \\ \hline
Mention & 257 & 84 & 56 & 145 \\ \hline
Concentration & 1333 & 427 & 537 & 705 \\ \hline
Reagent & 11142 & 3646 & 4004 & 5079 \\ \hline
Time & 2399 & 751 & 870 & 959 \\ \hline
Modifier & 4593 & 1554 & 1601 & 3476 \\ \hline
Device & 1752 & 618 & 468 & 911 \\ \hline
\end{tabular}
\caption{Frequency of various entity-types in different dataset splits.}
\label{tab:my-table}
\end{table*}

\subsection{Dataset}

Wet Lab Protocol (WLP) dataset consists of 615 unique protocols from 623 protocols released by \cite{kulkarni2018wetlab}. It excludes the following 8 duplicate protocols:

\indent \indent protocol 45 (duplicate of protocol 441)

\indent \indent protocol 459 (duplicate of protocol 310)

\indent \indent protocol 464 (duplicate of protocol 46)

\indent \indent protocol 480 (duplicate of protocol 473)

\indent \indent protocol 482 (duplicate of protocol 474)

\indent \indent protocol 483 (duplicate of protocol 475)

\indent \indent protocol 484 (duplicate of protocol 476)

\indent \indent protocol 621 (duplicate of protocol 570)\newline

After discarding the duplicate protocols, the remaining 615 unique protocols are re-annotated in brat by 3 annotators with 0.75 inter-annotator agreement, measured by span-level Krippendorff’s $ \alpha $. The annotators not only added the missing entity-relations but also rectified the inconsistencies.

\begin{table}[]
\centering
\begin{tabular}{|l|l|l|}
\hline
 & \#protocols & \#sentences \\ \hline
train\_data & 370 & 8444 \\ \hline
dev\_data & 122 & 2839 \\ \hline
test\_data & 123 & 2862 \\ \hline
test\_data\_2020 & 111 & 3562 \\ \hline
\end{tabular}
\caption{Statistics of different dataset splits.}
\label{tab:my-table}
\end{table}

The detailed class-wise statistics pertaining to each of the dataset splits provided in the task are shown in Table 1. Corresponding number of protocols and sentences are provided in Table 2. Here, train\_data denotes the training dataset, dev\_data denotes the validation dataset, test\_data denotes the test dataset and test\_data\_2020 denotes the surprise test dataset. The surprise dataset was not revealed before the evaluation window.

\begin{table}[]
\centering
\begin{tabular}{|l|l|l|}
\hline
 & Vocabulary & OOV (wrt ref) \\ \hline
train\_data & 7397 & - \\ \hline
dev\_data & 4082 & 1148 \\ \hline
test\_data & 3946 & 982 \\ \hline
test\_data\_2020 & 5718 & 2461 \\ \hline
\end{tabular}
\caption{Out-of-Vocabulary statistics.}
\label{tab:my-table}
\end{table}

Table 3 presents the total number of words, words absent in reference and words present in reference for each dataset. Reference varies according to the dataset being considered. For validation dataset and test dataset, training dataset is the reference. For surprise dataset, all data i.e. the union of training dataset, validation dataset and test dataset is considered as the reference. There is no reference in case of training dataset.

\subsection{System Settings}

\begin{table}[]
\centering
\begin{tabular}{|l|l|}
\hline
Hyperparameter & Value \\ \hline
Embedding & Flair + PubMedBERT \\ \hline
Final layer type & BiLSTM \\ \hline
Final layer hidden size & 512 \\ \hline
\# Final layers & 2 \\ \hline
CRF & \ding{51}  \\ \hline
Patience epochs & 3 \\ \hline
max epochs & 30 \\ \hline
Initial learning rate & 0.1 \\ \hline
Mini batch size & 32 \\ \hline
Merging scheme & SLE \\ \hline
\end{tabular}
\caption{System Settings for the final model.}
\label{tab:my-table}
\end{table}

\begin{table*}[]
\centering
\begin{tabular}{|l|l|l|l|l|l|l|l|}
\hline
\multirow{2}{*}{Base Model} & \multirow{2}{*}{Finetuning} & \multicolumn{3}{l|}{Final Layer} & \multirow{2}{*}{CRF} & \multirow{2}{*}{micro-F1} & \multirow{2}{*}{macro-F1} \\ \cline{3-5}
 &  & Type & \#layers & \#dim &  &  &  \\ \hline
Bert-base-cased & \ding{51} & Dense & 1 & - & \ding{55} & 78.78 & 71.06 \\ \hline
Bert-base-cased & \ding{55} & BiLSTM & 1 & 128 & \ding{51} & 80.56 & 73.41 \\ \hline
BioBERT & \ding{55} & BiLSTM & 1 & 128 & \ding{51} & 81.01 & 73.81 \\ \hline
PubmedBERT & \ding{55} & BiLSTM & 1 & 128 & \ding{51} & 81.36 & 73.67 \\ \hline
Flair & \ding{55} & BiLSTM & 1 & 128 & \ding{51} & 81.63 & 74.84 \\ \hline
PubmedBERT + Flair & \ding{55} & BiLSTM & 1 & 128 & \ding{51} & 81.71 & 75.22 \\ \hline
PubmedBERT + Flair & \ding{55} & BiLSTM & 2 & 374 & \ding{51} & 82.06 & 75.18 \\ \hline
PubmedBERT + Flair & \ding{55} & BiLSTM & 2 & 512 & \ding{51} & 82.28 & 75.57 \\ \hline
\end{tabular}
\caption{Results of experiments to identity the best architecture specification.}
\label{tab:my-table}
\end{table*}

While training individual models of our final ensemble, we rely on concatenated word representations from PubMedBERT and Flair. We train the BiLSTM-CRF based model with 3 BiLSTM layer each of hidden size 512 using a patience-based strategy. With this strategy, after every epoch of training, we compute the F1-score on validation split and if the metric doesn’t improve continuously for “patience” number of epochs, we reduce the learning rate by half. We ultimately stop the training when either the learning rate diminishes to 0.0001 or the epoch number reaches a maximum limit. We have utilised hugging-face\footnote{\url{https://huggingface.co/transformers/}} BERT APIs and Flair Framework\cite{akbik-etal-2019-flair} to train our model. We ran our experiments on a single NVIDIA V100 GPU. It took around 2.5 hours to train each individual model of our final submitted ensemble. Table 4 summarises the hyper-parameters which we employed to train our models.

\subsection{Evaluation Metrics}

Assuming that P and T represent the set of predicted and ground-truth entities for a particular word in the protocol text. Then, precision, recall and F1-score for the entity prediction of the considered word is defined as follows: 

\[ Precision =  \frac{\left | P \bigcap T \right |}{\left | P \right |} \] 
\[ Recall =  \frac{\left | P \bigcap T \right |}{\left | T \right |} \] 
\[ F1 =  \frac{2 \ast Precision \ast Recall }{Precision + Recall} \] 

There were two criteria for evaluation metrics in the task, partial match and exact match. In case of partial match, P intersection T will include all entities whose types match and boundaries match partially, i.e. there is some overlap in the boundaries. However, in case of exact match, for an entity to be included in the intersection set, it must have the same type as well as exact same boundaries. 

\subsection{Results and Error Analysis}

\begin{table}[]
\centering
\begin{tabular}{|l|l|l|}
\hline
\#ensembles & MajV & SLE \\ \hline
3 & 82.32 & 82.50 \\ \hline
5 & 82.52 & 82.68 \\ \hline
7 & 82.52 & 82.58 \\ \hline
9 & 82.55 & 82.64 \\ \hline
11 & 82.60 & 82.74 \\ \hline
\end{tabular}
\caption{micro-F1 on test-set after ensembling.}
\label{tab:my-table}
\end{table}

\begin{table}[]
\centering
\begin{tabular}{|l|l|l|}
\hline
P\_Label & T\_Label & Count \\ \hline
O & B-Modifier & 324 \\ \hline
B-Modifier & O & 287 \\ \hline
O & I-Modifier & 247 \\ \hline
B-Reagent & I-Reagent & 180 \\ \hline
I-Reagent & B-Reagent & 112 \\ \hline
B-Modifier & B-Reagent & 122 \\ \hline
O & B-Action & 137 \\ \hline
O & I-Reagent & 115 \\ \hline
O & I-Method & 112 \\ \hline
B-Action & O & 190 \\ \hline
\end{tabular}
\caption{Top-10 errors occurring in model predictions.}
\label{tab:my-table}
\end{table}

\begin{table}[]
\centering
\begin{tabular}{|l|l|l|}
\hline
 & Exact Match & Partial Match \\ \hline
Precision & 81.36 & 85.74 \\ \hline
Recall & 74.12 & 78.11 \\ \hline
Micro-F1 & 77.57 & 81.75 \\ \hline
\end{tabular}
\caption{Final results on surprise-test dataset.}
\label{tab:my-table}
\end{table}

Our approach involved working in two phases, first in which we experiment with different model architectures and the second in which we experiment with two output merging schemes. The results of our experiments in Phases 1 and 2 are summarised in Table 5 and 6 respectively. In Table 5, we present the micro-F1 and macro-F1 scores for different model architectures we experiment with by varying the base model, fine tuning implementation, type and specifications of final layer and CRF layer addition. Table 6 presents the micro-F1 scores on the test set when we experiment with the number of ensembles, i.e. on merging different number of prediction sets.

For our final submission to WNUT Shared Task-1, we employed an ensemble of eleven individual models. Each of these models was trained on a random train-validation split of original train + validation + test dataset. Our ensemble achieved a micro-F1 score of 0.8175 and 0.7757 for the partial and exact match of entity boundaries, respectively. We achieved highest micro-recall score among all the participating teams. In Table 7, we report the top-10 confusions which our model makes while assigning entity type to different words. Results of the final submission on surprise test set are summarised in Table 8. Upon close inspection of predicted outputs on test split, we identified the following error patterns in the model predictions:

\begin{itemize}
\item From Table 7, we can see that model dominantly gets confused while identifying the begin and intermediate tags for class Reagent. Upon inspection of the predictions, we identified that such errors were more common when the Reagent class in validation/test set was unseen in training examples. We can come up with a dictionary based approach to improve the precision of tags specifically for the Reagent class.
\item Modifier entity type modifies the semantics of some other entity type, so for a word to be Modifier or not is highly dependent on context and modified entity. But since our model fails to over-rely on context for recognition of certain entities, Modifier entity-type often gets confused with Other type. 
\item For the entities corresponding to numerical values like Concentration, Amount, Size  and Numerical, model often gets confused among such entities. The main reason which we suspect is that to classify these entities, the model should over-rely on context and not on the token corresponding to the entity itself. Since tokens can be shared across different classes. e.g. 1.5 ml microcentrifuge tube; Preds: B-Amount I-Amount B-Location I-Location; True Label: B-Size I-Size B-Location I-Location; 
\end{itemize}

\section{Conclusion and Future Work}

Through this paper, we showcased our approach to tackle the Shared Task 1 in EMNLP WNUT-2020 Workshop which involved Entity Recognition over Wet Lab Protocols. We solved the task in two phases. The first phase involved experimenting with different contextualised word embeddings like BERT and Flair, and a BiLSTM-CRF model to find the best performing model configuration for the problem at hand. In the second phase, we create an ensemble consisting of eleven BiLSTM-CRF models. We train individual models on randomly shuffled train-validation splits of the complete dataset. Also, we experiment with different merging techniques like Majority Voting and Structured Learning Ensemble (SLE). Our end solution achieved a micro F1-score of 0.8175 and 0.7757 in the partial and exact match categories, respectively. We were ranked first and second in partial and exact match categories respectively. In the future, we wish to explore the idea of employing rule-based approach to overcome the shortcomings of current solution.

\bibliographystyle{acl_natbib}
\bibliography{emnlp2020}

\begin{thebibliography}{21}
\expandafter\ifx\csname natexlab\endcsname\relax\def\natexlab#1{#1}\fi

\bibitem[{Adejo and Connolly(2017)}]{article}
Olugbenga Adejo and Thomas Connolly. 2017.
\newblock \href {https://doi.org/10.1108/JARHE-09-2017-0113} {Predicting
  student academic performance using multi-model heterogeneous ensemble
  approach}.
\newblock \emph{Journal of Applied Research in Higher Education}, 10:00--00.

\bibitem[{Akbik et~al.(2019)Akbik, Bergmann, Blythe, Rasul, Schweter, and
  Vollgraf}]{akbik-etal-2019-flair}
Alan Akbik, Tanja Bergmann, Duncan Blythe, Kashif Rasul, Stefan Schweter, and
  Roland Vollgraf. 2019.
\newblock \href {https://doi.org/10.18653/v1/N19-4010} {{FLAIR}: An easy-to-use
  framework for state-of-the-art {NLP}}.
\newblock In \emph{Proceedings of the 2019 Conference of the North {A}merican
  Chapter of the Association for Computational Linguistics (Demonstrations)},
  pages 54--59, Minneapolis, Minnesota. Association for Computational
  Linguistics.

\bibitem[{Ananthanarayanan and Thies(2010)}]{ananthanarayanan2010biocoder}
Vaishnavi Ananthanarayanan and William Thies. 2010.
\newblock Biocoder: A programming language for standardizing and automating
  biology protocols.
\newblock \emph{Journal of biological engineering}, 4(1):1--13.

\bibitem[{Bates et~al.(2017)Bates, Berliner, Lachoff, Jaschke, and
  Groban}]{bates2017wet}
Maxwell Bates, Aaron~J Berliner, Joe Lachoff, Paul~R Jaschke, and Eli~S Groban.
  2017.
\newblock Wet lab accelerator: a web-based application democratizing laboratory
  automation for synthetic biology.
\newblock \emph{ACS synthetic biology}, 6(1):167--171.

\bibitem[{Devlin et~al.(2018)Devlin, Chang, Lee, and
  Toutanova}]{devlin2018bert}
Jacob Devlin, Ming-Wei Chang, Kenton Lee, and Kristina Toutanova. 2018.
\newblock \href {http://arxiv.org/abs/1810.04805} {Bert: Pre-training of deep
  bidirectional transformers for language understanding}.

\bibitem[{Gu et~al.(2020)Gu, Tinn, Cheng, Lucas, Usuyama, Liu, Naumann, Gao,
  and Poon}]{pubmedbert}
Yu~Gu, Robert Tinn, Hao Cheng, Michael Lucas, Naoto Usuyama, Xiaodong Liu,
  Tristan Naumann, Jianfeng Gao, and Hoifung Poon. 2020.
\newblock \href {http://arxiv.org/abs/arXiv:2007.15779} {Domain-specific
  language model pretraining for biomedical natural language processing}.

\bibitem[{Hanafiah and Quix(2014)}]{10.1007/978-3-319-05476-6_12}
Novita Hanafiah and Christoph Quix. 2014.
\newblock Entity recognition in information extraction.
\newblock In \emph{Intelligent Information and Database Systems}, pages
  113--122, Cham. Springer International Publishing.

\bibitem[{Hochreiter and Schmidhuber(1997)}]{Hochreiter1997LongSM}
S.~Hochreiter and J.~Schmidhuber. 1997.
\newblock Long short-term memory.
\newblock \emph{Neural Computation}, 9:1735--1780.

\bibitem[{King et~al.(2009)King, Rowland, Oliver, Young, Aubrey, Byrne,
  Liakata, Markham, Pir, Soldatova et~al.}]{king2009automation}
Ross~D King, Jem Rowland, Stephen~G Oliver, Michael Young, Wayne Aubrey, Emma
  Byrne, Maria Liakata, Magdalena Markham, Pinar Pir, Larisa~N Soldatova,
  et~al. 2009.
\newblock The automation of science.
\newblock \emph{Science}, 324(5923):85--89.

\bibitem[{Kulkarni et~al.(2018)Kulkarni, Xu, Ritter, and
  Machiraju}]{kulkarni2018wetlab}
Chaitanya Kulkarni, Wei Xu, Alan Ritter, and Raghu Machiraju. 2018.
\newblock An annotated corpus for machine reading of instructions in wet lab
  protocols.
\newblock In \emph{Proceedings of the 2018 Conference of the North American
  Chapter of the Association for Computational Linguistics: Human Language
  Technologies (NAACL)}.

\bibitem[{Lee et~al.(2019)Lee, Yoon, Kim, Kim, Kim, So, and Kang}]{Lee_2019}
Jinhyuk Lee, Wonjin Yoon, Sungdong Kim, Donghyeon Kim, Sunkyu Kim, Chan~Ho So,
  and Jaewoo Kang. 2019.
\newblock \href {https://doi.org/10.1093/bioinformatics/btz682} {Biobert: a
  pre-trained biomedical language representation model for biomedical text
  mining}.
\newblock \emph{Bioinformatics}.

\bibitem[{Li et~al.(2018)Li, Sun, Han, and Li}]{li2018survey}
Jing Li, Aixin Sun, Jianglei Han, and Chenliang Li. 2018.
\newblock \href {http://arxiv.org/abs/1812.09449} {A survey on deep learning
  for named entity recognition}.

\bibitem[{Nguyen and Guo(2007)}]{10.1145/1273496.1273582}
Nam Nguyen and Yunsong Guo. 2007.
\newblock \href {https://doi.org/10.1145/1273496.1273582} {Comparisons of
  sequence labeling algorithms and extensions}.
\newblock In \emph{Proceedings of the 24th International Conference on Machine
  Learning}, ICML '07, page 681–688, New York, NY, USA. Association for
  Computing Machinery.

\bibitem[{Nozza et~al.(2020)Nozza, Bianchi, and Hovy}]{nozza2020mask}
Debora Nozza, Federico Bianchi, and Dirk Hovy. 2020.
\newblock \href {http://arxiv.org/abs/2003.02912} {What the [mask]? making
  sense of language-specific bert models}.

\bibitem[{Schuster and Paliwal(1997)}]{schuster1997bidirectional}
M.~Schuster and K.K. Paliwal. 1997.
\newblock \href {https://doi.org/10.1109/78.650093} {Bidirectional recurrent
  neural networks}.
\newblock \emph{Trans. Sig. Proc.}, 45(11):2673--2681.

\bibitem[{Shi and Demberg(2019)}]{shi-demberg-2019-next}
Wei Shi and Vera Demberg. 2019.
\newblock \href {https://doi.org/10.18653/v1/D19-1586} {Next sentence
  prediction helps implicit discourse relation classification within and across
  domains}.
\newblock In \emph{Proceedings of the 2019 Conference on Empirical Methods in
  Natural Language Processing and the 9th International Joint Conference on
  Natural Language Processing (EMNLP-IJCNLP)}, pages 5790--5796, Hong Kong,
  China. Association for Computational Linguistics.

\bibitem[{Sutton and McCallum(2010)}]{sutton2010introduction}
Charles Sutton and Andrew McCallum. 2010.
\newblock \href {http://arxiv.org/abs/1011.4088} {An introduction to
  conditional random fields}.

\bibitem[{Tabassum et~al.(2020)Tabassum, Xu, and Ritter}]{tabassum2020wlp}
Jeniya Tabassum, Wei Xu, and Alan Ritter. 2020.
\newblock {WNUT-2020 Task 1: Extracting Entities and Relations from Wet Lab
  Protocols}.
\newblock In \emph{Proceedings of EMNLP 2020 Workshop on Noisy User-generated
  Text (WNUT)}.

\bibitem[{Vasilev et~al.(2011)Vasilev, Liu, Haddock, Bhatia, Adler, Yaman,
  Beal, Babb, Weiss, and Densmore}]{vasilev}
Viktor Vasilev, Chenkai Liu, Traci Haddock, Swapnil Bhatia, Aaron Adler, Fusun
  Yaman, Jacob Beal, Jonathan Babb, Ron Weiss, and Douglas Densmore. 2011.
\newblock A software stack for specification and robotic execution of protocols
  for synthetic biological engineering.

\bibitem[{Vaswani et~al.(2017)Vaswani, Shazeer, Parmar, Uszkoreit, Jones,
  Gomez, Kaiser, and Polosukhin}]{vaswani2017attention}
Ashish Vaswani, Noam Shazeer, Niki Parmar, Jakob Uszkoreit, Llion Jones,
  Aidan~N. Gomez, Lukasz Kaiser, and Illia Polosukhin. 2017.
\newblock \href {http://arxiv.org/abs/1706.03762} {Attention is all you need}.

\bibitem[{Yadav and Bethard(2018)}]{yadav2019survey}
Vikas Yadav and Steven Bethard. 2018.
\newblock \href {https://www.aclweb.org/anthology/C18-1182} {A survey on recent
  advances in named entity recognition from deep learning models}.
\newblock In \emph{Proceedings of the 27th International Conference on
  Computational Linguistics}, pages 2145--2158, Santa Fe, New Mexico, USA.
  Association for Computational Linguistics.

\end{thebibliography}

\end{document}